\documentclass[conference]{IEEEtran}
\usepackage{times}

\usepackage[numbers]{natbib}
\usepackage{multicol}
\usepackage[bookmarks=true]{hyperref}

\usepackage{graphicx}
\usepackage{amsmath}
\usepackage{amssymb}

\usepackage{subfigure}
\usepackage{xcolor}
\usepackage{footnote}
\usepackage{makecell}
\usepackage{multirow, booktabs}
\graphicspath{{./figures/}{./figures/draw/}}

\newcommand\red[1]{{\color{black}#1}}

\newcommand{\junk}[1]{}

\newcommand{\method}[1]{Deep-6DPose}
\def\R{{\mathbb R}}

\newcommand\norm[1]{\left\lVert#1\right\rVert}

\begin{document}

\title{\method{}: Recovering 6D Object Pose \\from a Single RGB Image}
\author{Thanh-Toan Do, Ming Cai, Trung Pham, Ian Reid\\
The University of Adelaide, Australia}



%

\maketitle

\begin{abstract}
Detecting objects and their 6D poses from only RGB images is an important task for many robotic applications. While deep learning methods have made significant progress in visual object detection and segmentation, the object pose estimation task is still challenging. In this paper, we introduce an end-to-end deep learning framework, named \textit{\method{}}, that jointly detects, segments, and most importantly recovers 6D poses of object instances from a single RGB image. In particular, we extend the recent state-of-the-art instance segmentation network \emph{Mask R-CNN} with a novel pose estimation branch to directly regress 6D object poses without any post-refinements. Our key technical contribution is the decoupling of pose parameters into translation and rotation so that the rotation can be regressed via a Lie algebra representation. The resulting pose regression loss is differential and unconstrained, making the training tractable. The experiments on two standard pose benchmarking datasets show that our proposed approach compares favorably with the state-of-the-art RGB-based multi-stage pose estimation methods. Importantly, due to the end-to-end architecture, \textit{\method{}} is considerably faster than competing multi-stage methods, offers an inference speed of 10 fps that is well suited for robotic applications. 

\end{abstract}

\IEEEpeerreviewmaketitle

\section{Introduction}
Detecting objects and their 6D poses (3D location and orientation) is an important task for many robotic applications including object manipulations (e.g., pick and place), parts assembly, to name a few. In clutter environments, objects must be first detected before their poses can be estimated. Accurate object segmentation is also important, especially when objects are occluded. Our goal is to develop a deep learning framework that jointly detects, segments,  and most importantly recovers 6D poses of object instances from a single RGB image. While the first two tasks are getting more and more mature thanks to the power of deep learning, the 6D pose estimation problem remains a challenging problem.

Traditional object pose estimation methods are mainly based on the matching of hand-crafted local features (e.g. SIFT~\cite{SIFT_Lowe}). However, these local feature matching based approaches are only suitable for richly textured objects. For poorly textured objects, template-based matching~\cite{ACCV12,DBLP:conf/eccv/TejaniTKK14,DBLP:conf/iccv/Rios-CabreraT13} or dense feature learning approaches are usually used~\cite{ECCV14,ICCV15,CVPR17,CVPR17_2}. However, the template-based methods are usually sensitive to illuminations and occlusions.  
The feature learning approaches~\cite{ECCV14,ICCV15,CVPR17,CVPR17_2} have shown better performances against template-based methods, but they suffer from several disadvantages, i.e., 
they require a time-consuming multi-stage processing for learning dense features, generating coarse pose hypotheses, and refining the coarse poses. 



With the rising of deep learning, especially Convolutional Neural Networks (CNN), the object classification~\cite{krizhevsky2012imagenet}, object detection~\cite{Fast-RCNN,Faster-RCNN}, and recently object instance segmentation~\cite{Mask-RCNN,affnet} tasks have achieved remarkable  improvements. However, the application of CNN to 6D object pose estimation problem is still limited. 
Recently, there are few works~\cite{SSD-6D,BB8,posecnn} which apply deep learning for 6D object pose estimation. These methods, however, are not end-to-end or only estimate a coarse object poses. They require further post-refinements to improve the accuracy, which linearly increases the running time, w.r.t. the number of detected objects. 

\junk{
The key idea for recent large improvements in object detection and object instance segmentation problems~\cite{Faster-RCNN,Mask-RCNN} is the development of a CNN architecture, i.e., Region Proposal Network (RPN)~\cite{Faster-RCNN}. RPN is actually a CNN which is trained to produce multiple object (bounding boxes) proposals in an image at different shapes and sizes.  
Faster R-CNN~\cite{Faster-RCNN} further refines and classify bounding boxes produced by RPN using additional fully connected layers. 
The recent work Mask R-CNN~\cite{Mask-RCNN} goes beyond Faster-RCNN, i.e., it performs  binary segmentation in each bounding box produced by RPN. 
}

Recently, Mask R-CNN~\cite{Mask-RCNN} achieves state-of-the-art results in the instance segmentation problem. The key component of Mask R-CNN is a Region Proposal Network (RPN)~\cite{Faster-RCNN}, which predicts multiple object (bounding boxes) proposals in an image at different shapes and sizes. 
Mask R-CNN further segments instances inside bounding boxes produced from RPN by using additional convolutional layers. 

Inspired by the impressive results of Mask R-CNN for object instance segmentation, 
we are motivated to find the answer for the question that, \textit{can we exploit the merits of RPN to not only segment but also recover the poses of object instances in a single RGB image, in an end-to-end fashion?}
To this end, we design a network which simultaneously detects\footnote{The detection means the prediction of both bounding boxes and class labels.}, segments, and also recovers 6D poses of object instances from a single RGB image. In particular, we propose a \method{} network, which goes beyond Mask R-CNN by adding a novel branch for regressing the poses for the object instances inside bounding boxes produced by RPN. The proposed pose branch is parallel with the detection and segmentation branches. 

Our main contribution is a novel object pose regressor, where the network regresses translation and rotation parameters seperately. Cares must be taken when regressing 3D rotation matrices as not all $3\times3$ matrices are valid rotation matrices. To work around, we resort to the Lie algebra associated with the $SO(3)$ Lie group for our 3D rotation representation. Compared to other representations such as quaternion or orthonormal matrix, Lie algebra is an optimal choice as it is less parammeters and unconstrained, thus making the training process easier. Although the Lie algebra representation has been widely used in geometry-based robot vision problems~\cite{DBLP:conf/iros/Agrawal06,DBLP:conf/icra/RosGSPL13}, to our best knowledge, this is the first work which successfully uses the Lie algebra in a CNN for regressing 6D object poses.


Different from recent deep learning-based 6D pose estimation methods which are not end-to-end trainable~\cite{BB8} or only predict a rough pose followed by a pose refinement step~\cite{SSD-6D,BB8,posecnn}, the proposed \method{} is a single deep learning architecture. It takes a RGB image as input and directly outputs 6D object poses without any pose post-refinements. Additionally, our system also returns segmentation masks of object instances. The experimental results show that \method{} is competitive or outperforms the state-of-the-art methods on standard datasets. Furthermore, \method{} is simple and elegant, allows the inference at the speed of $\textrm{10 fps}$, which is several times faster than many existing methods.

The remainder of this paper is organized as follows. Section~\ref{sec:related} presents related works. Section~\ref{sec:method} details the proposed \method{}. Section~\ref{sec:exp} evaluates and compares \method{} to the state-of-the-art 6D object pose estimation methods. Section~\ref{sec:conl} concludes the paper. 

\section{Related Work}
\label{sec:related}
In this section, we first review the 6D object pose estimation methods.  We then brief the main design of the recent methods which are based on RPN for object detection and segmentation.

\textbf{Classical approaches.} The topic of pose estimation has great attention in the past few years. For objects with rich of texture, sparse feature matching approaches have been shown good accuracy~\cite{DBLP:conf/clor/GordonL06,SIFT_Lowe,DBLP:conf/icra/MartinezCS10}. Recently, researchers have put more focus on poor texture or texture-less objects. The most traditional approaches for poor texture objects are to use object templates~\cite{ACCV12,DBLP:conf/eccv/TejaniTKK14,DBLP:conf/iccv/Rios-CabreraT13}. The most notable work belonging to this category is LINEMOD~\cite{ACCV12} which is based on stable gradient and normal features. However, LINEMOD is designed to work with RGBD images. Furthermore, template-based approaches are sensitive to the lighting and occlusion. 

\textbf{Feature learning approach.} Recent 6D pose estimation researches have relied on feature learning for dealing with insufficient texture objects~\cite{ECCV14,ICCV15,CVPR17,CVPR17_2}. In~\cite{ECCV14,ICCV15}, the authors show that the dense feature learning approach outperforms matching approach. The basic  design of \cite{ECCV14,ICCV15,CVPR17,CVPR17_2} is a time-consuming multi-stage pipeline, i.e., a random forest is used for jointly learning the object categories for pixels 
and the coordinates of pixels w.r.t. object coordinate systems (known as object coordinates). A set of pose hypotheses is generated by using the outputs of the forest and the depth channel of the input image. 
An energy function is defined on the generated pose hypotheses to select hypotheses. The selected pose hypotheses are further refined to obtain the final pose. Note that the pipelines in those works heavily depend on the depth channel. The depth information is required in both pose hypothesis generation and refinement. The work~\cite{CVPR16} also follows a multi-stage approach as~\cite{ECCV14,ICCV15} but is designed to work with RGB inputs. 
In order to deal with the missing depth information, the distribution of object coordinates is approximated as a mixture model when generating pose hypotheses. 

\junk{
The disadvantage of feature learning approaches~\cite{ECCV14,ICCV15,CVPR16,CVPR17_2} is that the generation of pose hypotheses uses only local information, i.e., only three or four pixels are used to generate a hypothesis. As result, this may generate bad hypotheses because it does not consider a global context over the whole object. Furthermore, by requiring multiple processing steps, those approaches are are time-consuming, making them unsuitable for real-time applications. 
}

\textbf{CNN-based approach.} In recent years, CNN has been applied for 6D pose problem, firstly for camera pose~\cite{DBLP:conf/iccv/KendallGC15,kendall2017posenet}, and recently for object pose~\cite{ICCV15,CVPR17_2,SSD-6D,BB8,posecnn,yolo-6D}. 

In~\cite{DBLP:conf/iccv/KendallGC15,kendall2017posenet}, the authors train CNNs to directly regress 6D camera pose from a single RGB image. The camera pose estimation task is arguably easier than the object pose estimation task, because to estimate object pose, it also requires accurate detection and classification of the object, while these steps are not typically needed for camera pose. 

In~\cite{ICCV15}, the authors use a CNN in their object pose estimation system. However,  
the CNN is only used as a probabilistic model to learn to compare the learned information (produced by a random forest) and the rendered image. The CNN outputs an energy value for selecting pose hypotheses which are further refined. 
The work in~\cite{CVPR17_2} improves over~\cite{ICCV15} by using a CNN and reinforcement learning for joint selecting and refining pose hypotheses. 

In SSD-6D~\cite{SSD-6D}, the authors extend SSD detection framework~\cite{SSD} to 3D detection and 3D rotation estimation. The authors decompose 3D rotation space into discrete viewpoints and in-plane rotations. They then treat the rotation estimation as a classification problem. 
However, to get the good results, it is required to manually find an appropriate sampling for the rotation space. Furthermore, the approach SSD-6D does not directly output the translation, i.e., to estimate  the translation, for each object, an offline stage is required to precomputes bounding boxes w.r.t. all possible sampled rotations. This precomputed information is used together with the estimated bounding box and rotation to estimate the 3D translation. 
In the recent technical report~\cite{posecnn}, the authors propose a network, dubbed PoseCNN, which jointly segments objects and estimates the rotation and the distance of segmented objects to camera. However, by relying on a semantic segmentation approach (which is a FCN~\cite{FCN}) to localize objects, it may be difficult for PoseCNN to deal with input image which contains multiple instances of an object. 
Both SSD-6D and PoseCNN also require further pose refinement steps to improve the accuracy. 

In BB8~\cite{BB8}, the authors propose a cascade of multiple CNNs for object pose estimation task. A segmentation network is firstly applied to the input image to localize objects. Another CNN is then used to predict 2D projections of the corners of the 3D bounding boxes around objects. The 6D pose is estimated for the correspondences between the projected 2D coordinates and the 3D ground control points of 
bounding box corners using a PnP algorithm. Finally, a CNN per object is trained to refine the pose. By using multiple separated CNNs, BB8 is not end-to-end and is time-consuming for the inference. 
Similar to~\cite{BB8}, in the recent technical report~\cite{yolo-6D}, the authors extend YOLO object detection network~\cite{yolo9000} to predict 2D projections of the corners of the 3D bounding boxes around objects. Given the projected 2D coordinates and the 3D ground control points of bounding box corners, a PnP algorithm is further used to estimate the 6D object pose.  

\textbf{RPN-based detection and segmentation.} One of key components in the recent successful object detection method  Faster R-CNN~\cite{Faster-RCNN} and instance segmentation method Mask R-CNN~\cite{Mask-RCNN} is the Region Proposal Network --- RPN. The core idea of RPN is to dense sampling the whole input image by many overlap bounding boxes at different shapes and sizes. The network is trained to produce multiple object proposals (also known as Region of Interest --- RoI). This design of RPN allows to smoothly search over different scales of feature maps. 
\junk{
For each RoI, a fixed-size small feature map (e.g., $7\times7$) is pooled from the image feature map using the RoIPool layer~\cite{RCNN} or RoIAlign layer~\cite{Mask-RCNN}. These layers work by dividing the RoI into a regular grid and then max-pooling the feature map values in each grid cell. In Faster R-CNN, the outputs of the RoIPool layer are used to refine the RoI coordinates and to classify the RoI label. In Mask-RCNN, the outputs of the RoIAlign layer are used not only for refining and recognizing the RoI but also for segmenting the object inside the RoI. 
}
Faster R-CNN~\cite{Faster-RCNN} further refines and classifies RoIs with additional fully connected layers, 
while Mask R-CNN~\cite{Mask-RCNN} further improves over Fast R-CNN by segmenting instances inside RoIs with additional convolutional layers. 

In this paper, we go beyond Mask RCNN. In particular, depart from the backbone of Mask R-CNN, we propose a novel head branch which takes RoIs from RPN as inputs to regress the 6D object poses and is parallel with the existing branches. This results a novel end-to-end architecture which is not only detecting, segmenting but also directly recovering the 6D poses of object instances from a single RGB image. 


\section{Method}
\label{sec:method}
Our goal is to simultaneously detect, segment, and estimate the 6D poses of object instances in the input image. 
Mask R-CNN performs well for the first two tasks, except the 6D pose estimation. In order to achieve a complete system, we propose a novel branch which takes RoIs from RPN as inputs and outputs the 6D poses of the instances inside the RoIs. 
Although the concept is simple, the additional 6D pose is distinct from the other branches. It requires an effective way to represent the 6D pose and a careful design of the loss function. 
In this paper, we represent a pose by a 4-dimensional vector, in which the first three elements represent the Lie algebra associated with the rotation matrix of the pose; the last element represents the $z$ component of the translation vector of the pose. Given the predicted $z$ component and the predicted bounding box from the box regression branch, we use projective property to recover the full translation vector. The architecture of \method{} is shown in Figure~\ref{fig:overview}.

\subsection{\method{}}

\begin{figure*}[!t] 
\centering   				
\includegraphics[scale=0.4]{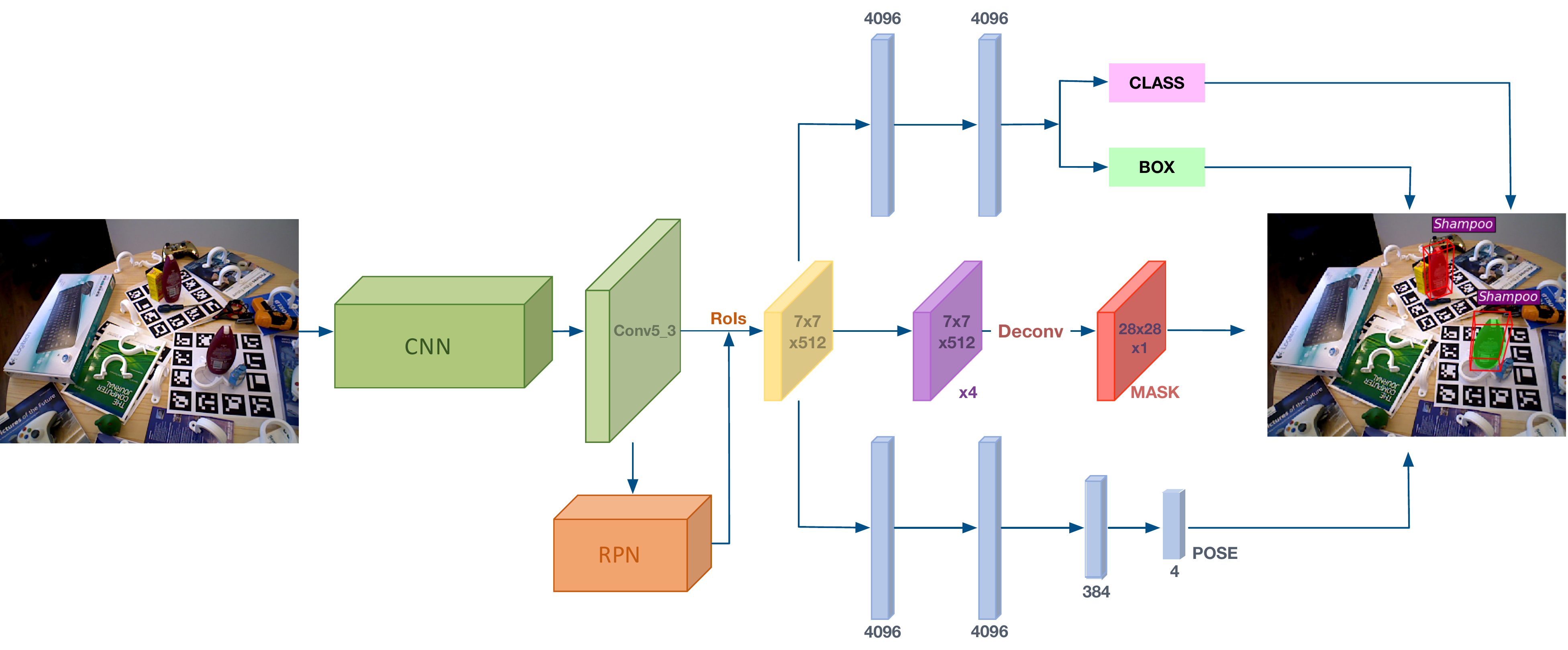}    
    \caption{An overview of \method{} framework. {From left to right:} The input to \method{} is a RGB image. A deep CNN backbone (i.e., VGG) is used to extract features over the whole image. The RPN is attached on the last convolutional layer of VGG (i.e., $conv5\_3$) and outputs RoIs. For each RoI, the corresponding features from the feature map $conv5\_3$ are extracted and pooled into a fixed size $7\times 7$. The pooled features are used as inputs for $4$ head branches. For the box regression and classification heads, we follow Mask-RCNN~\cite{Mask-RCNN}. The segmentation head is \textit{adapted} from~\cite{Mask-RCNN}, i.e., four $3\times3$ consecutive convolutional layers (denoted as `$\times4$') are used. The ReLu is used after each convolutional layer. A deconvolutional layer is used to upsample the feature map to $28\times28$ which is the segmentation mask. The proposed pose head consists of four fully connected layers. The ReLu is used after each of the first three fully connected layers. The last fully connected layer outputs four numbers which represent for the pose. As shown on the right image, the network outputs the detected instances (with classes, i.e., Shampoo), the predicted segmentation masks (different object instances are shown with different colors) and the predicted 6D poses for detected instances (shown with 3D boxes). 
}
    \label{fig:overview} 
\end{figure*}

Let us first briefly recap Mask R-CNN~\cite{Mask-RCNN}. Mask R-CNN consists of two main components. The first component is a RPN~\cite{Faster-RCNN} which produces candidate RoIs. The second component extracts features from each candidate RoI using RoIAlign layer~\cite{Mask-RCNN} and performs the classification, the bounding box regression, and the segmentation. We refer readers to~\cite{Mask-RCNN} for details of Mask-RCNN. 

\method{} also consists of two main components. The first component is also a RPN. In the second component, in \textit{parallel} to the existing branches of Mask R-CNN, \method{} also outputs a 6D pose for the objects inside RoIs. 

\paragraph{Pose representation}
An important task when designing the pose branch is the representation space of the output poses. We learn the translation in the Euclidean space. In stead of predicting full translation vector, our network is trained to regress the $z$ component only. The reason is that when projecting a 3D object model into a 2D image, two translation vectors with the same $z$ and the different $x$ and $y$ components may produce two objects which have very similar appearance and scale in 2D image (at different positions in the image -- in the extreme case of parallel projection, there is no difference at all). This causes difficulty for the network to predict the $x$ and $y$ components by using only appearance information as input. However, the object size and the scale of its textures in a 2D image provide strong cues about the $z$-coordinate. 
This projective property allows the network to learn the $z$ component of the translation using the 2D object appearance only. Given the $z$ component, it is used together with predicted bounding box, which is outputted by the bounding box regression branch, to fully recover the translation. 
The detail of this recovering process is presented in the following sections.  

Representing the rotation part of the pose is more complicated than the translation part. Euler angles are intuitive due to the explicit meaning of parameters. 
However, the Euler angles wrap around at 2$\pi$ radians, i.e., having multiple values representing the same angle. This causes difficulty in learning a uni-modal scalar regression task.  
Furthermore, the Euler angles-based representation suffers from the well-studied problem of gimbal lock~\cite{pose}. Another alternative, the use of $3\times3$ orthonormal matrix is over-parametrised, and creates the problem of enforcing the orthogonality constraint when training the network through back-propagation. A final common representation is the unit length 4-dimensional quaternion. 
One of the downsides of quaternion representation is its norm should be unit. This constraint may harm the optimization~\cite{DBLP:conf/iccv/KendallGC15}. 
 
In this work, we use the Lie algebra $so(3)$ associated with the Lie group $SO(3)$ (which is space of 3D rotation matrices) as our rotation representation. 
The Lie algebra $so(3)$  is known as the tangent space at the identity element of the Lie group $SO(3)$. 
We choose the Lie algebra $so(3)$ to represent the rotation because an arbitrary element of $so(3)$ admits a skew-symmetric matrix representation parameterized by a vector in $\R^3$ which is continuous and smooth.
This means that the network needs to regress only three scalar numbers for a rotation, without any constraints. 
To our best knowledge, this paper is the first one which uses Lie algebra for representing rotations in training a deep network for 6D object pose estimation task. 

During training, we map the groundtruths of rotation matrices to their associated elements in $so(3)$ by the closed form Rodrigues logarithm mapping~\cite{log-exp-map}. 
The mapped values are used as regression targets when learning to predict the rotation. 

In summary, the pose branch is trained to regress a 4-dimensional vector, in which the first three elements represent rotation part and the last element represents the $z$ component of the  translation part of the pose. 

\paragraph{Multi-task loss function}
In order to train the network, we define a multi-task loss to jointly train the bounding
box class, the bounding box position, the segmentation, and the pose of the object inside the box. Formally, the loss function is defined as follows
\begin{equation}
 L=\alpha_1L_{cls} + \alpha_2L_{box} + \alpha_3L_{mask} + \alpha_4L_{pose} 
 \label{eq:allloss}
\end{equation}
  The classification loss $L_{cls}$, the bounding box regression loss $L_{box}$, and the segmentation loss $L_{mask}$ are defined similar as~\cite{Mask-RCNN}, which are \textit{softmax} loss, \textit{smooth L1} loss, and \textit{binary cross entropy} loss, respectively. The $\alpha_1, \alpha_2, \alpha_3, \alpha_4$  coefficients are scale factors to control the important of each loss during training. 

The pose branch outputs 4 numbers for each RoI, which represents the Lie algebra for the rotation and $z$ component of the translation. It is worth noting that in our design, the output of pose branch is class-agnostic, 
but the class-specific counterpart (i.e., with a $4C$-dimensional output vector in which $C$ is the number of classes) is also applicable. 
The pose regression loss $L_{pose}$ is defined as follows
\begin{equation}
L_{pose} = \norm{r-\hat{r}}_p + \beta\norm{t_z - \hat{t}_z}_p
\label{eq:poseloss}
\end{equation}
where $r$ and $\hat{r}$ are two 3-dimensional vectors representing the regressed rotation and the groundtruth rotation, respectively; $t_z$ and $\hat{t}_z$ are two scalars representing the regressed $z$ and the groundtruth $z$ of the translation; $p$ is a distance norm; $\beta$ is a scale factor to control the rotation and translation regression errors. 

\paragraph{Network architecture}
Figure~\ref{fig:overview} shows the schematic overview of \method{}. We differentiate two parts of the network, i.e., the backbone and the head branches. The backbone is used to extract features over the whole image and is shared between head branches. There are four head branches corresponding to the four different tasks, i.e., the bounding box regression, the bounding box classification, the segmentation, and the 6D pose estimation for the object inside the box. 
For the backbone, we follow Faster R-CNN~\cite{Faster-RCNN} which uses VGG~\cite{SimonyanZ14} together with a RPN attached on the last convolutional layer of VGG (i.e., $conv5\_3$). For each output RoI of RPN, a fixed-size $7\times7$ feature map is pooled from the $conv5\_3$ feature map using the RoIAlign layer~\cite{Mask-RCNN}. This pooled feature map is used as input for head branches. For the network heads of the bounding box regression and classification, we closely follow the Mask R-CNN~\cite{Mask-RCNN}. 
For the segmentation head, we adapt from Mask R-CNN. In our design, four $3\times3$ consecutive convolutional layers (denoted as `$\times4$' in Figure~\ref{fig:overview}) are used after the pooled feature map. The ReLu is used after each convolutional layer. A deconvolutional layer is used to upsample the feature map to $28\times28$ which is the segmentation mask. It is worth noting that for segmentation head, we use the class-agnostic design, i.e., this branch outputs a single mask, regardless of class. We empirically found that this design reduces the model complexity and the inference time, while it is nearly effective as the class-specific design. This observation is consistent with the observation in Mask R-CNN~\cite{Mask-RCNN}. 

In order to adapt the shared features to the specific pose estimation task, the pose head branch consists of a sequence of 4 fully connected layers in which the number of outputs are $4096 \to 4096 \to 384 \to 4$.  The ReLU is used after each fully layer, except for the last layer. This is because the regressing targets (i.e., the groundtruths) contain both negative and positive values. We note that our pose head has a simple structure. More complex design may have potential improvements. 

\subsection{Training and inference}
\paragraph{Training}
We implement \method{} using Caffe deep learning library~\cite{DBLP:conf/mm/JiaSDKLGGD14}. The input to our network is a RGB image with the size $480 \times 640$. 
The RPN outputs RoIs at different sizes and shapes. We use 5 scales and 3 aspect ratios, resulting 15 anchors in the RPN. The 5 scales are $16 \times 16$, $32\times32$, $64 \times 64$, $128\times128$ and $256 \times 256$; the 3 aspect ratios are $2:1$, $1:1$, $1:2$. This design allows the network to detect small objects. 

The $\alpha_1, \alpha_2, \alpha_3$, and $\alpha_4$ in (\ref{eq:allloss}) are empirically set to 1, 1, 2, 2, respectively. The values of $\beta$ in (\ref{eq:poseloss}) is empirically set to 1.5. 
An important choice for the pose loss (\ref{eq:poseloss}) is the regression norm $p$. Typically, deep learning models use $p=1$ or $p=2$. With the datasets used in this work, we found that $p=1$ give better results and hence is used in our experiments. 

We train the network in an end-to-end manner using stochastic gradient descent with $0.9$ momentum and $0.0005$ weight decay. The network is trained on a Titan X GPU for $350k$ iterations. Each mini batch has $1$ image. 
The learning rate is set to $0.001$ for the first $150k$ iterations and then decreased by 10 for the remaining iterations. The top $2000$ RoIs from RPN (with a ratio of 1:3 of positive to negative) are subsequently used for computing the multi-task loss. A RoI is considered positive if it has an intersection over union (IoU) with a groundtruth box of at least 0.5 and negative otherwise. The losses $L_{mask}$ and $L_{pose}$ are defined for only positive RoIs. 

\paragraph{Inference}
At the test phase, we run a forward pass on the input image.  The top $1,000$ RoIs produced by the RPN are selected and fed into the box regression and classification branches, followed by non-maximum suppression~\cite{Fast-RCNN}. Based on the outputs of the classification branch, we select the outputted boxes from the regression branch that have classification scores higher than a certain threshold (i.e., $0.9$) as the detection results. The segmentation branch and the pose branch are then applied on the detected boxes, which output segmentation masks and the 6D poses for the objects inside the boxes. 

\paragraph{From the 4-dimensional regressed pose to the full 6D pose}
Given the predicted Lie algebra, i.e., the first three elements of the predicted 4-dimensional vector from pose branch, we use the exponential Rodrigues mapping~\cite{log-exp-map} to map it to the corresponding rotation matrix.  
In order to recover the full translation, we rely on the predicted $z$ component ($t_z$ -- the last element of the 4-dimensional predicted vector) and the predicted bounding box coordinates to compute two missing components $t_x$ and $t_y$. We assume that the bounding box center (in 2D image) is the projected point of the 3D object center (the origin of the object coordinate system). Under this assumption, using the 3D-2D projection formulation, we compute $t_x$ and $t_y$ as follows
\begin{equation}
t_x = \frac{(u_0 - c_x)t_z}{f_x}
\end{equation}
\begin{equation}
t_y = \frac{(v_0 - c_x)t_z}{f_y}
\end{equation}
where $u_0$, $v_0$ are the bounding box center in 2D image, and the matrix $[f_x, 0, c_x; 0, f_y, c_y; 0, 0, 1]$ is the known intrinsic camera calibration matrix.

\section{Experiments}
\label{sec:exp}
We evaluate \method{} on two widely used datasets, i.e., the single object pose dataset LINEMOD provided by Hinterstoisser et al.~\cite{ACCV12} and the multiple object instance pose dataset provided by Tejani et al.~\cite{DBLP:conf/eccv/TejaniTKK14}.  
We also compare \method{} to the state-of-the-art methods for 6D object pose estimation from RGB images~\cite{CVPR16,BB8,SSD-6D}.

\junk{When evaluating on the dataset of Hinterstoisser et al.~\cite{ACCV12}, we mainly target to compare our work with the recent state-of-the-art 6D pose estimation method from Brachmann et al.~\cite{CVPR16} because that work also estimates 6D pose from a single RGB input~\cite{CVPR16}. For reference purposes, we also mention the results of LINE2D~\cite{LINE2D}, which is a template-based approach\footnote{LINE2D~\cite{LINE2D} is originally proposed for object detection. It is then extended for 6D pose estimation by~\cite{CVPR16}.}. It is because the method~\cite{CVPR16} has already improves over LINE2D~\cite{LINE2D}.}

\textbf{Metric:}
Different 6D pose measures have been proposed in the past. In order to evaluate the recovered poses, we use the \textit{standard} metrics used in~\cite{CVPR16,BB8}. To measure pose error in 2D, we project the 3D object model into the image using the groundtruth pose and the estimated pose. The estimated pose is accepted if the IoU between two project boxes is higher than 0.5. This metric is called as \textit{2D-pose} metric. To measure the pose error in 3D, the $5cm5^\circ$ and \textit{$\textrm{ADD}$} metrics is used. 
In $5cm5^\circ$ metric,  an estimated pose is accepted if it is within $5cm$ translational error and $5^\circ$ angular error of the ground truth pose.
In $\textrm{ADD}$ metric, an estimated pose is accepted if the average distance between transformed model point clouds by the groundtruth pose and the estimated pose is smaller than $10\%$ of the object's diameter. We also provide the $F1$ score of the detection and segmentation results. A detection / segmentation is accepted if its IoU with the groundtruth box / segmentation mask is higher than a threshold. We report results with the widely used thresholds, i.e., $0.5$ and $0.9$~\cite{Faster-RCNN,Mask-RCNN}.
\subsection{Single object pose estimation}
\label{exp:hoi}

\begin{table*}[!t]
\vspace{0.5cm}
   \centering
   \footnotesize
    \begin{center}
    \begin{tabular}{c| c c c c c c c c c c c c c c} 
&Ape &Bvise &Cam &Can &Cat &Driller &Duck &Box &Glue &Holep &Iron &Lamp &Phone &Average\\  
\Xhline{0.75pt}
	&\multicolumn{14}{c}{\textbf{IoU 0.5}} \\ 
Detection &99.8 &100 &99.7 &100 &99.5 &100 &99.8 &99.5 &99.2 &99.0 &100 &99.8 &100 &99.7\\ 
Segmentation &99.5 &99.8 &99.7 &100 &99.1 &100 &99.4 &99.5 &99.0 &98.6 &99.2 &99.4 &99.7 &99.4
\\  \hline

	&\multicolumn{14}{c}{\textbf{IoU 0.9}} \\
Detection 	 &85.4 &91.7 &93.3 &93.6 &89.3 &87.5 &86.3 &94.2 &81.1 &93.2 &92.5 &91.3 &90.8 &90.0\\ 
Segmentation &80.6 &57.0 &91.4 &62.5 &52.1 &74.6 &81.2 &91.9 &73.3 &84.6 &90.3 &85.0 &84.6 &77.6\\ \hline
    \end{tabular}
    \end{center}
    \vspace{-0.2cm}
    \caption{F1 score for 2D detection and segmentation of \method{} on LINEMOD dataset~\cite{ACCV12} for single object.} 
    \label{tab:2D_det_seg}
\end{table*}

\begin{table*}[!t]
   \centering
   \footnotesize
   \begin{center}
    \begin{tabular}{c| c c c c c c c c c c c c c c} 
&Ape &Bvise &Cam &Can &Cat &Driller &Duck &Box &Glue &Holep &Iron &Lamp &Phone &Average\\  
\Xhline{1pt}
	&\multicolumn{14}{c}{\textbf{2D-pose metric}}\\ 
\method{} &99.8 &100 &99.7 &100 &99.2 &100 &99.8 &99.0 &97.1 &98.0 &99.7 &99.8 &99.1 &99.3\\ 
Brachmann\cite{CVPR16} &98.2 &97.9 &96.9 &97.9 &98.0 &98.6 &97.4 &98.4 &96.6 &95.2 &99.2 &97.1 &96.0 &97.5\\
SSD-6D\cite{SSD-6D} &- &- &- &- &- &- &- &- &- &- &- &- &- &99.4\\

   	\hline
   	   	   	&\multicolumn{14}{c}{$\mathbf{5cm5^\circ}$ \textbf{metric}} \\
\method{} &57.8 &72.9 &75.6 &70.1 &70.3 &72.9 &67.1 &68.4 &64.6 &70.4 &60.7 &70.9 &69.7 &68.5\\ 
Brachmann\cite{CVPR16} &34.4 &40.6 &30.5 &48.4 &34.6 &54.5 &22.0 &57.1 &23.6 &47.3 &58.7 &49.3 &26.8 &40.6\\
BB8\cite{BB8} &80.2 &81.5 &60.0 &76.8 &79.9 &69.6 &53.2 &81.3 &54.0 &73.1 &61.1 &67.5 &58.6 &69.0\\
   	\hline   	
   	   	&\multicolumn{14}{c}{$\mathbf{ADD}$ \textbf{metric}} \\
\method{} &38.8 &71.2 &52.5 &86.1 &66.2 &82.3 &32.5 &79.4 &63.7 &56.4 &65.1 &89.4 &65.0 &65.2\\ 
Brachmann\cite{CVPR16}  &33.2 &64.8 &38.4 &62.9 &42.7 &61.9 &30.2 &49.9 &31.2 &52.8 &80.0 &67.0 &38.1 &50.2\\
BB8\cite{BB8} &40.4 &91.8 &55.7 &64.1 &62.6 &74.4 &44.3 &57.8 &41.2 &67.2 &84.7 &76.5 &54.0 &62.7\\
SSD-6D\cite{SSD-6D} &- &- &- &- &- &- &- &- &- &- &- &- &- &76.3\\
   	\hline   	
    \end{tabular}
    \end{center}
   \vspace{-0.2cm}
    \caption{Pose estimation accuracy on the LINEMOD dataset~\cite{ACCV12} for single object.}
    \label{tab:pose} 
\end{table*}

\begin{figure*}[!t]
\vspace{0.1cm}
\centering
\subfigure{
       \includegraphics[scale=0.17]{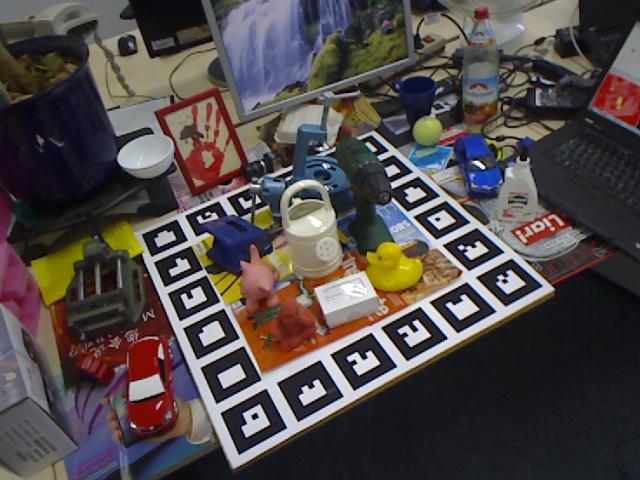}
}
\subfigure{
       \includegraphics[scale=0.17]{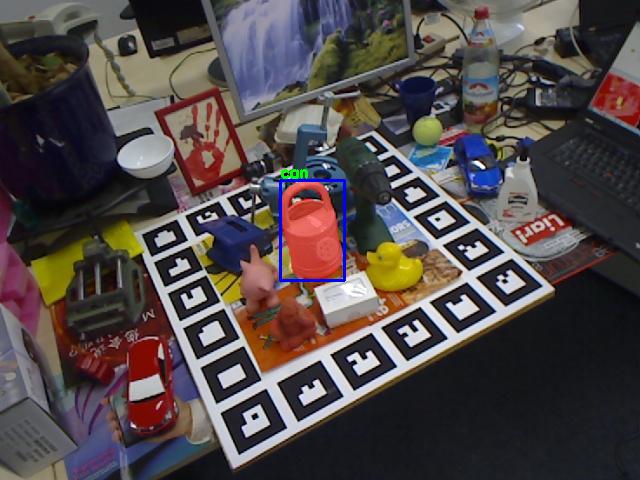} 
}
\subfigure{
       \includegraphics[scale=0.17]{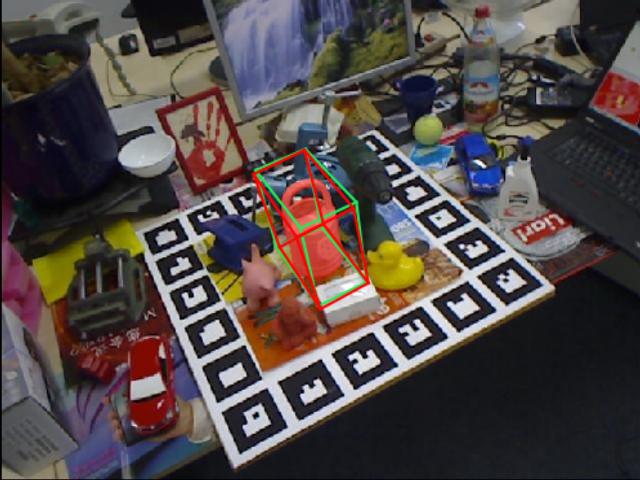} 
}
\\
\vspace{0.1cm}
\subfigure{
       \includegraphics[scale=0.17]{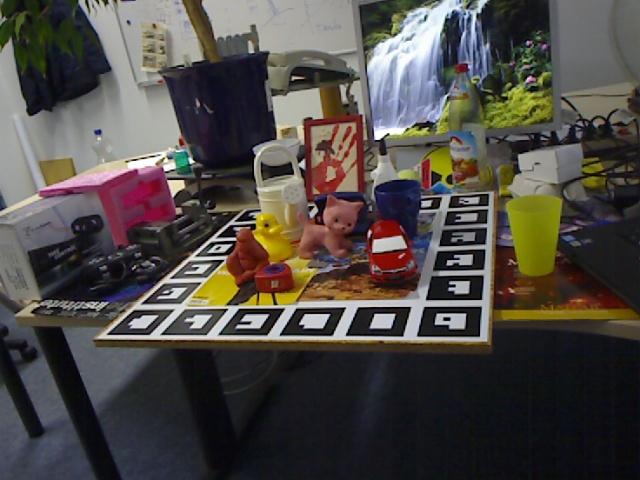}
}
\subfigure{
       \includegraphics[scale=0.17]{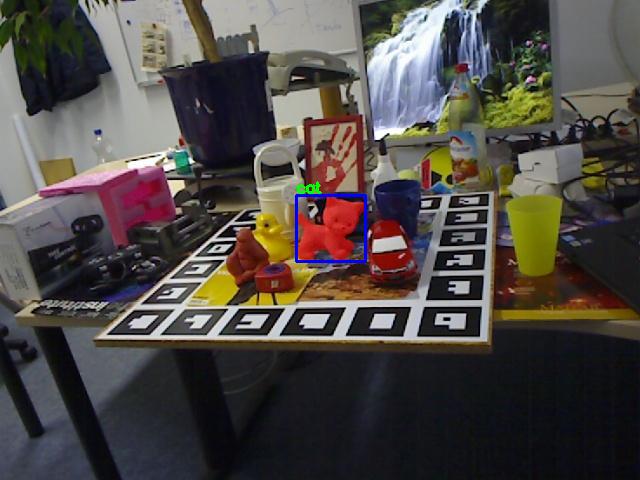} 
}
\subfigure{
       \includegraphics[scale=0.17]{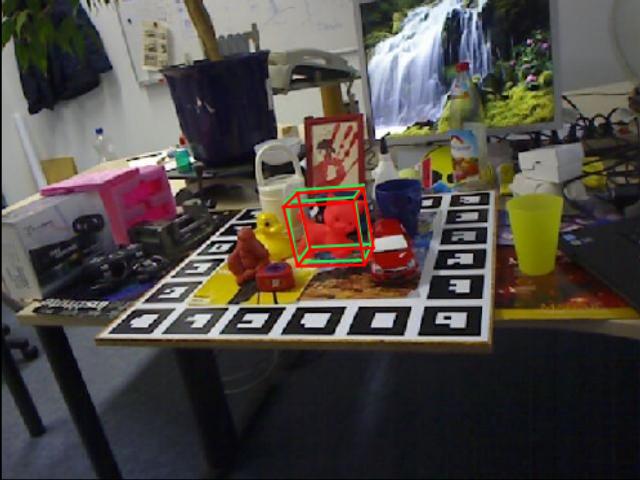} 
}
\\
\vspace{0.1cm}
\subfigure{
       \includegraphics[scale=0.17]{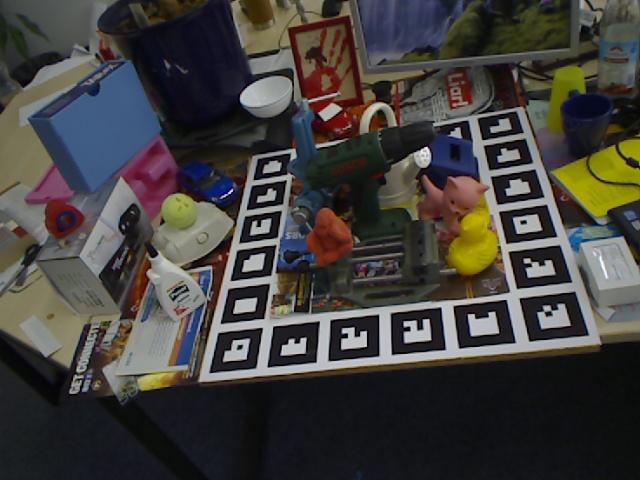} 
}
\subfigure{
       \includegraphics[scale=0.17]{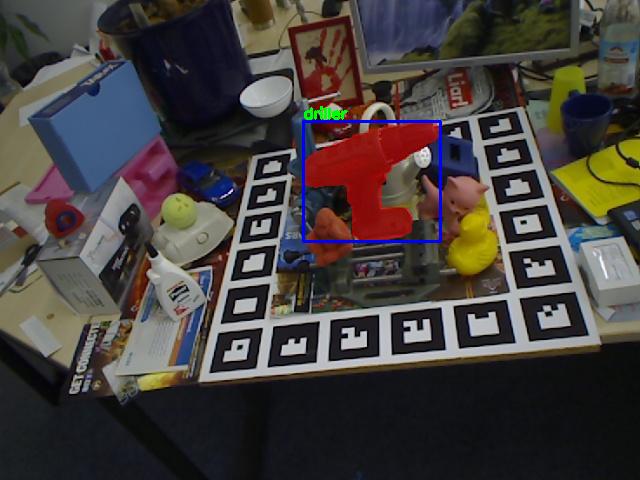} 
}
\subfigure{
       \includegraphics[scale=0.17]{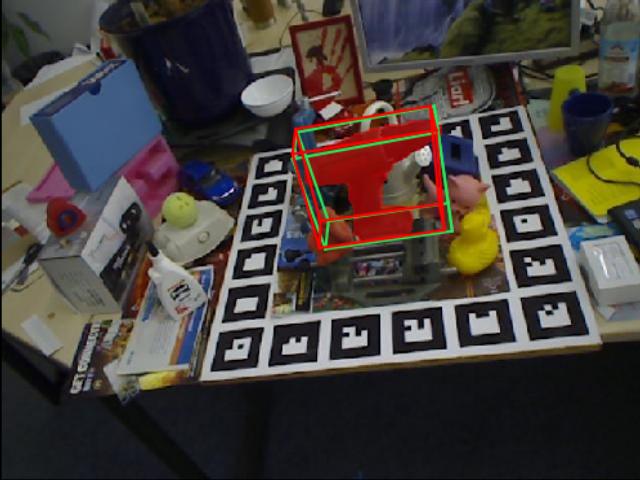} 
}
\caption[]{Qualitative results for single object pose estimation on the LINEMOD dataset~\cite{ACCV12}. From left to right: (i) original images, (ii) the predicted 2D bounding boxes, classes, and segmentations, (iii) 6D poses in which the green boxes are the groundtruth poses and the red boxes are the predicted poses. Best view in color.}
\label{fig:singleobj}
\vspace{-0.3cm}
\end{figure*}
In~\cite{ACCV12}, the authors publish LINEMOD, a RGBD dataset,  which has become a \textit{de facto} standard benchmark for 6D pose estimation. The dataset contains poorly textured objects in a cluttered scene. We only use the RGB images to evaluate our method. The dataset contains 15 object sequences. For fair comparison to~\cite{CVPR16,BB8,SSD-6D}, we evaluate our method on the $13$ object sequences for which the 3D models are available. 
The images in each object sequence contain multiple objects, however, only one object is annotated with the groundtruth class label, bounding box, and 6D pose. The camera intrinsic matrix is also provided with the dataset. Using the given groundtruth 6D poses, the object models, and the camera matrix, we are able to compute the groundtruth segmentation mask for the annotated objects. 
We follow the evaluation protocol in~\cite{CVPR16,BB8} which uses RGB images from the object sequences for training and testing. 
For each object sequence, 
\red{we randomly select $30\%$ of the images for training and validation. The remaining images serve as the test set.}   
A complication when using this dataset for training arises because  
not all objects in each image are annotated, i.e., only one object is annotated per sequence, even though multiple objects are present. This is problematic for training a detection/segmentation network such as~\cite{Faster-RCNN,Mask-RCNN} because the training may be confused, e.g. slow or fail to converge, if an object is annotated as foreground in some images and as background other images. 
Hence, we preprocess the training images as follows. For each object sequence, we use the RefineNet~\cite{Lin:2017:RefineNet}, a state-of-the-art semantic segmentation algorithm, to train a semantic segmentation model. The trained model is applied on all training images in other sequences. The predicted masks in other sequences are then filtered out, so that the appearance of the objects without annotated information does not hinder the training. 

\junk{
\paragraph{Evaluation protocol} 
In the evaluation protocol of~\cite{CVPR16}, the authors assume that for each testing image, the object sequence which the testing image belongs to is known. Hence they can select the learned features corresponding to the known object class for computing the pose. It helps to improve the accuracy. 
We do not require the object sequence to be known in our evaluation. From the detection results which have already filtered by a global threshold, i.e. 0.9, for each object class, we keep only one detection with the highest classification score. This allows us to access the 2D detection and segmentation performance. We count a 2D detection / segmentation to be correct if its IoU with the groundtruth box / segmentation mask is higher than a threshold. We report the detection and segmentation results with IoU thresholds 0.5 and 0.9. In order to evaluate the recovered poses, we follow the metrics used in~\cite{CVPR16}. To measure pose error in 2D, we project the 3D object model into the image using the groundtruth pose and the estimated pose. The estimated pose is accepted if the IoU between two project boxes is higher than 0.5. This metric is called as \textit{2D pose}. To measure the pose error in 3D, the $5cm5^\circ$ metric is used. An estimated pose is accepted if it is within $5cm$ translational error and $5^\circ$ angular error of the ground truth pose. \junk{The angular distance between two rotation matrices $P$ and $Q$ is computed as follows 
\begin{equation}
\theta = \arccos \frac{tr(R)-1}{2}
\end{equation}
where $R=PQ^T$.}
}

\textbf{Results:}
Table~\ref{tab:2D_det_seg} presents the 2D detection and segmentation results of \method{}. At an IoU 0.5, the results show that both detection and segmentation achieve nearly perfect scores for all object categories. This reconfirms the effective design of Faster R-CNN~\cite{Faster-RCNN} and Mask R-CNN~\cite{Mask-RCNN} for object detection and instance segmentation. 
When increasing the IoU to 0.9, both detection and segmentation accuracy  significantly decrease. The more decreasing is observed for the segmentation, i.e., the dropping around \red{10\%} and \red{22\%} for detection and segmentation, respectively. 

We put our interest on the pose estimation results, which is the main focus of this work. 
Table~\ref{tab:pose} presents the comparative pose estimation accuracy between \method{} and the state-of-the-art works of Brachmann et al.~\cite{CVPR16}, BB8~\cite{BB8}, SSD-6D~\cite{SSD-6D} which also use RGB images as inputs to predict the poses. Under \textit{2D-pose} metric, \method{} is comparable to SSD-6D, while outperforms over \cite{CVPR16} around \red{2\%}.
Under $5cm5^\circ$ metric, \method{} is slightly lower than BB8, while it significantly outperforms~\cite{CVPR16}, i.e., around \red{28\%}. Under $\textrm{ADD}$ metric, \method{} outperforms BB8 \red{2.5\%}. 
The results of \method{} are also more stable than BB8~\cite{BB8}, e.g., under $5cm5^\circ$ metric, the standard deviations in the accuracy of \method{} and BB8~\cite{BB8} are \red{5.0} and \red{10.6}, respectively. 
The Table~\ref{tab:pose} also shows that both \method{} and BB8~\cite{BB8} are worse than SSD-6D. However, it is worth noting that SSD-6D~\cite{SSD-6D} does not use images from the object sequences for training. 
The authors~\cite{SSD-6D} perform a discrete sampling over \textit{whole} rotation space and use the known 3D object models to generate synthetic images used for training. By this way, the training data of SSD-6D is able to cover more rotation space than~\cite{CVPR16}, BB8~\cite{BB8}, and~\method{}. 
Furthermore, SSD-6D also further uses an ICP-based refinement to improve the accuracy. Different from SSD-6D, \method{} directly outputs the pose without any post-processing. 
Figure~\ref{fig:singleobj} shows some qualitative results of \method{} for single object pose estimation on the LINEMOD dataset. 


\subsection{Multiple object instance pose estimation}
\label{exp:tejani}
In~\cite{DBLP:conf/eccv/TejaniTKK14}, Tejani et al. publish a dataset consisting of six object sequences in which the images in each sequence contain multiple instances of the same object with different levels of occlusion. Each object instance is provided with the groundtruth class label, bounding box, and 6D pose. Using the given groundtruth 6D poses, the object models, and the known camera matrix, we are able to compute the groundtruth segmentation masks for object instances. 
We use the RGB images provided by the dataset for training and testing. \red{We randomly split $30\%$ images in each sequence for training and validation. The remaining images serve as the test set.}

\junk{
\paragraph{Evaluation protocol} 
Because there are multiple instances of an object in an image, we do not constraint the maximum number of the detection results. We accept all detections whose the classification scores are higher than a threshold. The threshold is fixed to $0.9$ for all sequences. A 2D detection / segmentation is accepted if its IoU with the groundtruth box / segmentation mask is higher than a threshold. We report the results with IoU thresholds $0.5$ and $0.9$. 
In order to evaluate the recovered poses, we use {2D pose} and $5cm5^\circ$ metrics as mentioned in the first dataset. 
Because there are multiple object instances in each image, follow the original work~\cite{DBLP:conf/eccv/TejaniTKK14}, we report results with $F1$ score, which is the harmonic mean of precision and recall.
}

\begin{table*}[!t]
   \footnotesize
   \begin{center}
    \begin{tabular}{c| c c c c c c c} 
&Camera &Coffee &Joystick &Juice &Milk &Shampoo &Average\\  
\Xhline{0.75pt}
	&\multicolumn{7}{c}{\textbf{IoU 0.5}} \\ 
Detection &99.8 &100 &99.8 &99.2 &99.7 &99.5 &99.6 \\ 
Segmentation &99.8 &99.7 &99.6 &99.0 &99.3 &99.5 &99.4 \\ \hline
	&\multicolumn{7}{c}{\textbf{IoU 0.9}} \\ 
Detection     &88.3 &97.2 &97.7 &89.4 &83.5 &82.4 &89.7 \\ 
Segmentation  &81.7 &95.0 &95.5 &86.9 &77.8 &74.7 &85.2 \\ \hline
    \end{tabular}
    \end{center}
  \vspace{-0.2cm}
    \caption{F1 score for 2D detection and segmentation of \method{} on the dataset of Tejani et al.~\cite{DBLP:conf/eccv/TejaniTKK14} for multiple object instances.} 

   \label{tab:2D_det_seg_tejani}
\end{table*}
\begin{table*}[!t]
\vspace{-0.2cm}
   \centering
   \footnotesize
   \begin{center}
    \begin{tabular}{c| c c c c c c c} 
&Camera &Coffee &Joystick &Juice &Milk &Shampoo &Average\\  \Xhline{1pt}
	&\multicolumn{7}{c}{\textbf{2D pose metric}}\\ 
\method{}	    	&99.2 &100 &99.6 &98.4 &99.5 &99.1 &99.3 \\ 
SSD-6D\cite{SSD-6D} &-    &-   &-    &-    &-    &-    &98.8 \\
   	\hline
   	&\multicolumn{7}{c}{$\mathbf{5cm5^\circ}$ \textbf{metric}}\\ 
\method{}   &76.5 &18.7 &60.2 &85.6 &73.5 &72.4 &64.5 \\ \hline
   	&\multicolumn{7}{c}{$\mathbf{ADD}$ \textbf{metric}}\\ 
\method{}   &80.4 &35.4 &27.5 &81.2 &71.6 &75.8 &62.0 \\ 

   	\hline 	
    \end{tabular}
    \end{center}
    \vspace{-0.2cm}
    \caption{Pose estimation accuracy on the dataset of Tejani et al.~\cite{DBLP:conf/eccv/TejaniTKK14} for multiple object instances.} 
    \label{tab:pose_tejani}
\end{table*}
\begin{figure*}[!t]
\centering
\vspace{-0.3cm}
\subfigure{
       \includegraphics[scale=0.17]{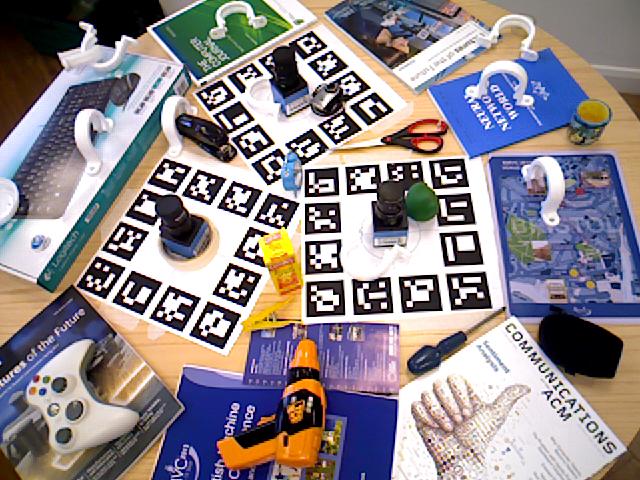}
}
\subfigure{
       \includegraphics[scale=0.17]{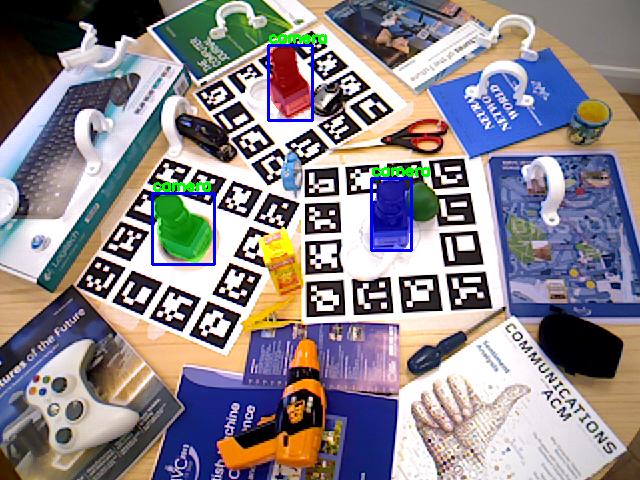} 
}
\subfigure{
       \includegraphics[scale=0.17]{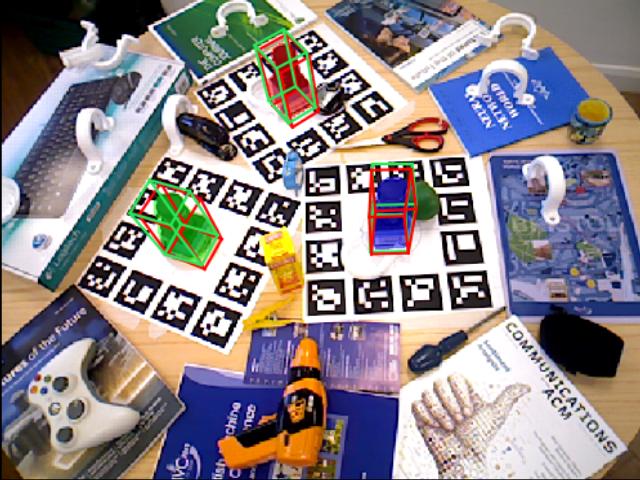} 

}\\
\vspace{-0.2cm}
\subfigure{
       \includegraphics[scale=0.17]{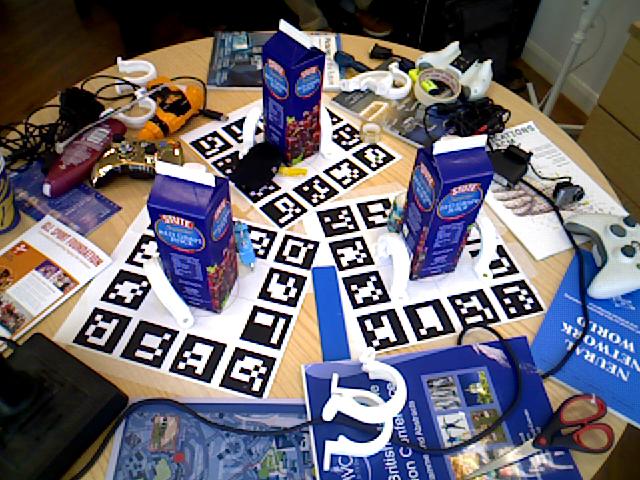}
}
\subfigure{
       \includegraphics[scale=0.17]{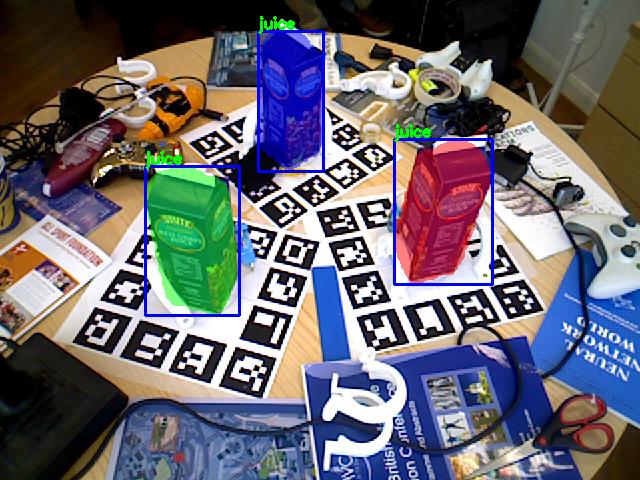} 
}
\subfigure{
       \includegraphics[scale=0.17]{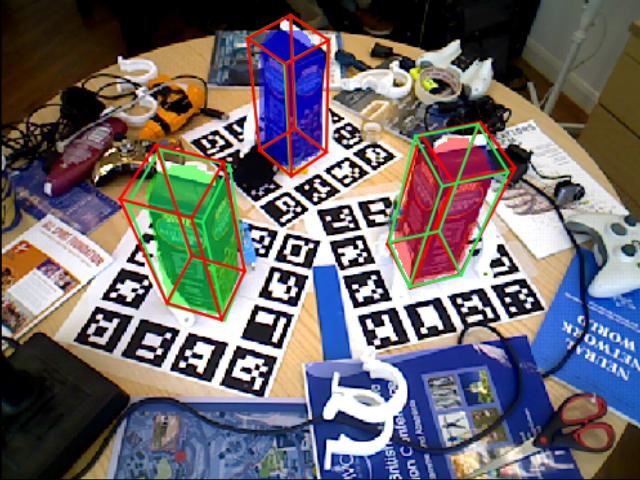} 
}
\\
\vspace{-0.2cm}
\subfigure{
       \includegraphics[scale=0.17]{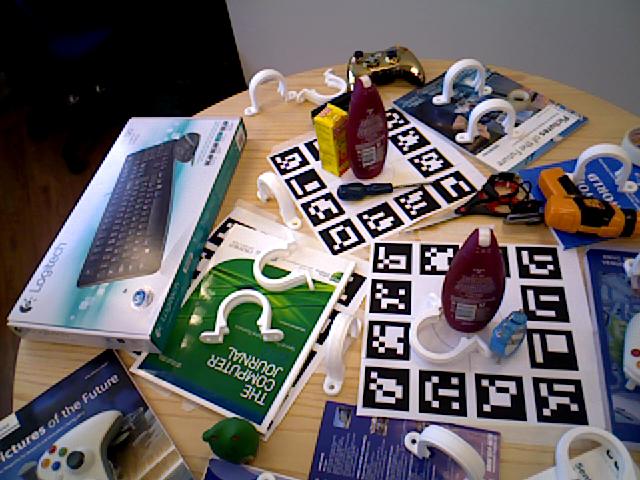}
}
\subfigure{
        \includegraphics[scale=0.17]{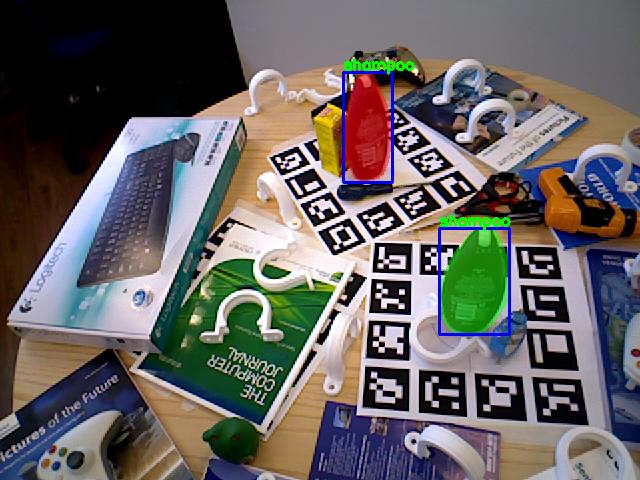} 
}
\subfigure{
       \includegraphics[scale=0.17]{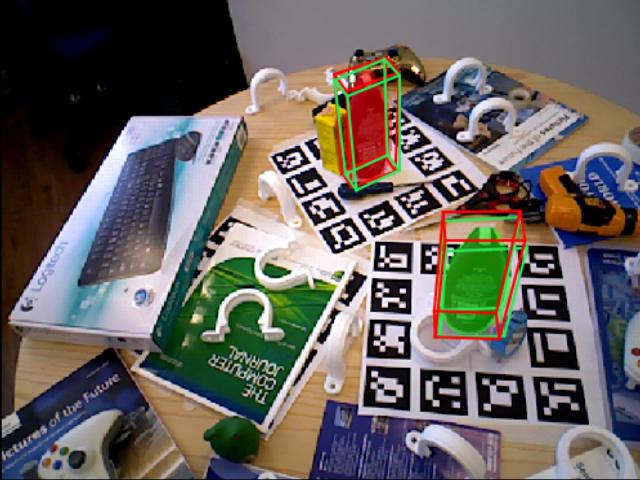} 
}
\caption[]{Qualitative results for pose estimation on the multiple object instance dataset of Tejani et al.~\cite{DBLP:conf/eccv/TejaniTKK14}. From left to right: (i) the original images, (ii) the predicted 2D bounding boxes, classes, and segmentations (different instances are shown with different colors), (iii) 6D poses in which the green boxes are the groundtruth poses and the red boxes are the predicted poses. Best view in color.}
\label{fig:singleobj_tenaji}
\vspace{-0.3cm}
\end{figure*}

\begin{figure}[!t]
\centering
\subfigure{
       \includegraphics[scale=0.15]{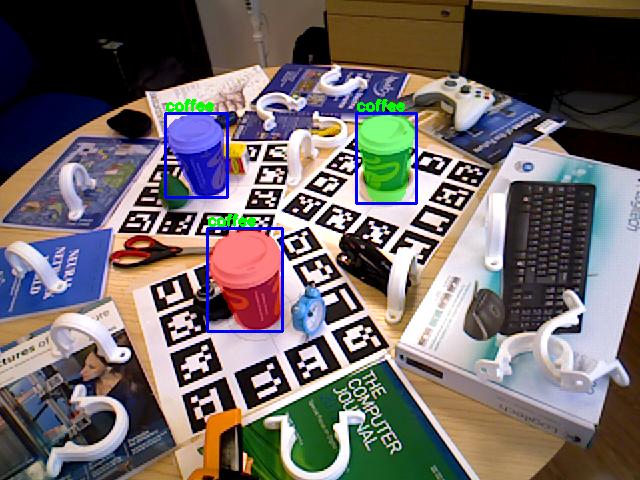}
}
\subfigure{
       \includegraphics[scale=0.15]{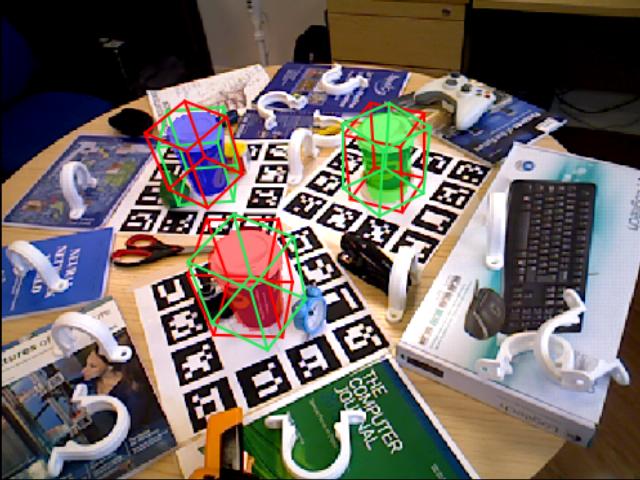} 
}
\caption[]{One fail case on the Coffee sequence in the dataset of Tejani et al.~\cite{DBLP:conf/eccv/TejaniTKK14}. Although the network can correctly predict the bounding box, the estimated 6D is incorrect due to the rotational symmetry of the object. We can see that the incorrect rotations are mainly in the Yaw axis.}
\label{fig:fail_tejani}
\end{figure}
\textbf{Results:} 
The 2D detection and segmentation results are presented in Table~\ref{tab:2D_det_seg_tejani}. 
At an IoU $0.5$, \method{} achieves nearly perfect scores. When increasing the IoU to $0.9$, the average accuracy decreases around \red{10\%} and \red{14\%} for detection and segmentation, respectively. We found that the Shampoo category has the most drop. It is caused by its flat shape, e.g., at some certain poses, the projected 2D images only contain a small side edge of the object, resulting the drop of scores at high IoU.

The accuracy of the recovered poses is reported in Table~\ref{tab:pose_tejani}. Under \textit{2D-pose} metric, both \method{} and SSD-6D~\cite{SSD-6D} achieve mostly perfect scores.  
 Under $5cm5^\circ$ and $\textrm{ADD}$ metrics, on the average, \method{} achieves \red{64.5\%} and \red{62.0\%} accuracy, respectively. We note that no previous works reports the standard $5cm5^\circ$ and $\textrm{ADD}$ metrics on this dataset using only RGB images. 
Figure~\ref{fig:singleobj_tenaji} shows some qualitative results for the predicted bounding boxes, classes, segmentations, and 6D poses for multiple object instances. 


We also found that \method{} is not very robust to nearly rotationally symmetric objects. That is because if an object is rotational symmetry in $z$ axis, any rotation of the 3D object in the Yaw angle will produce the same object appearance in the 2D image. This makes confusion for the network to predict the rotation using only appearance information. 
As shown in Table~\ref{tab:pose_tejani}, the Coffee sequence, which is nearly rotational symmetry in both shape and texture, has very low $5cm5^\circ$ score. Figure~\ref{fig:fail_tejani} shows a failure case for this object sequence.

\subsection{Timing}
When testing on the LINEMODE dataset~\cite{ACCV12}, Brachman {{et al.}}~\cite{CVPR16} reports a running time around $0.45s$ per image. \method{} is several times faster than their method, i.e., its end-to-end architecture allows the inference at around $0.1s$ per image on a Titan X GPU. SSD-6D~\cite{SSD-6D} and BB8~\cite{BB8} report the running time around $0.1s$ and $0.3s$ per image, respectively. \method{} is comparable to SSD-6D in the inference speed, while it is around 3 times faster than BB8. It is worth noting that due to using post-refinement, the inference time of SSD-6D and BB8 may be increased when the input image contains multiple object instances. We note that although \method{} is fast, its parameters are still not optimized for the speed.  
 As shown in the recent study~\cite{DBLP:journals/corr/HuangRSZKFFWSG016}, better trade-off between speed and accuracy may be achieved by carefully selecting parameters, e.g., varying the number of proposals after RPN, image sizes, which is beyond scope of the current work.

\section{Conclusion}
\label{sec:conl}
In this paper, we propose \method{}, a deep learning approach for jointly detecting,  segmenting, and most importantly recovering 6D poses of object instances from an single RGB image. \method{} is end-to-end trainable and can directly output estimated poses without any post-refinements. The novelty design is at a new pose head branch, which uses the Lie algebra to represent the rotation space. \method{} compares favorably with the state-of-the-art RGB-based 6D object pose estimation methods. Furthermore, \method{} also allows a fast inference which is around 10 fps. An interesting future work is to improve the network for handling  with rotationally symmetric objects. 


\bibliographystyle{plainnat}
\bibliography{ref}
\end{document}